\documentclass[10pt,twocolumn,letterpaper]{article}

\usepackage{cvpr}              

\usepackage{graphicx}
\usepackage{amsmath}
\usepackage{amssymb}
\usepackage{booktabs}

\usepackage{multirow}
\usepackage{bbding}

\usepackage[pagebackref,breaklinks,colorlinks]{hyperref}

\usepackage[capitalize]{cleveref}
\crefname{section}{Sec.}{Secs.}
\Crefname{section}{Section}{Sections}
\Crefname{table}{Table}{Tables}
\crefname{table}{Tab.}{Tabs.}

\def\cvprPaperID{3399} 
\def\confName{CVPR}
\def\confYear{2022}

\begin{document}

\title{Exploring Complicated Search Spaces with Interleaving-Free Sampling}

\author{Yunjie Tian\textsuperscript{1}, Lingxi Xie\textsuperscript{2}, Jiemin Fang\textsuperscript{3}, Jianbin Jiao\textsuperscript{1},\\
Qixiang Ye\textsuperscript{1}, Qi Tian\textsuperscript{2}\\
\textsuperscript{1}University of Chinese Academy of Sciences\quad \textsuperscript{2}Huawei Inc.,\\
\textsuperscript{3}Huazhong University of Science and Technology\quad \\
{\tt\small tianyunjie19@mails.ucas.ac.cn}\quad{\tt\small 198808xc@gmail.com}\quad \\
{\tt\small jaminfong@hust.edu.cn}\quad{\tt\small jiaojb@ucas.ac.cn}\quad 
{\tt\small qxye@ucas.ac.cn}\quad {\tt\small tian.qi1@huawei.com}
}
\maketitle

\begin{abstract}

The existing neural architecture search algorithms are mostly working on search spaces with short-distance connections. We argue that such designs, though safe and stable, obstacles the search algorithms from exploring more complicated scenarios. In this paper, we build the search algorithm upon a complicated search space with long-distance connections, and show that existing weight-sharing search algorithms mostly fail due to the existence of \textbf{interleaved connections}. Based on the observation, we present a simple yet effective algorithm named \textbf{IF-NAS}, where we perform a periodic sampling strategy to construct different sub-networks during the search procedure, avoiding the interleaved connections to emerge in any of them. In the proposed search space, IF-NAS outperform both random sampling and previous weight-sharing search algorithms by a significant margin. IF-NAS also generalizes to the micro cell-based spaces which are much easier. Our research emphasizes the importance of macro structure and we look forward to further efforts along this direction.
\end{abstract}

\section{Introduction}
\label{section:introduction}

\begin{figure}
\vskip -0.01in
\centering
\includegraphics[width=7.5cm]{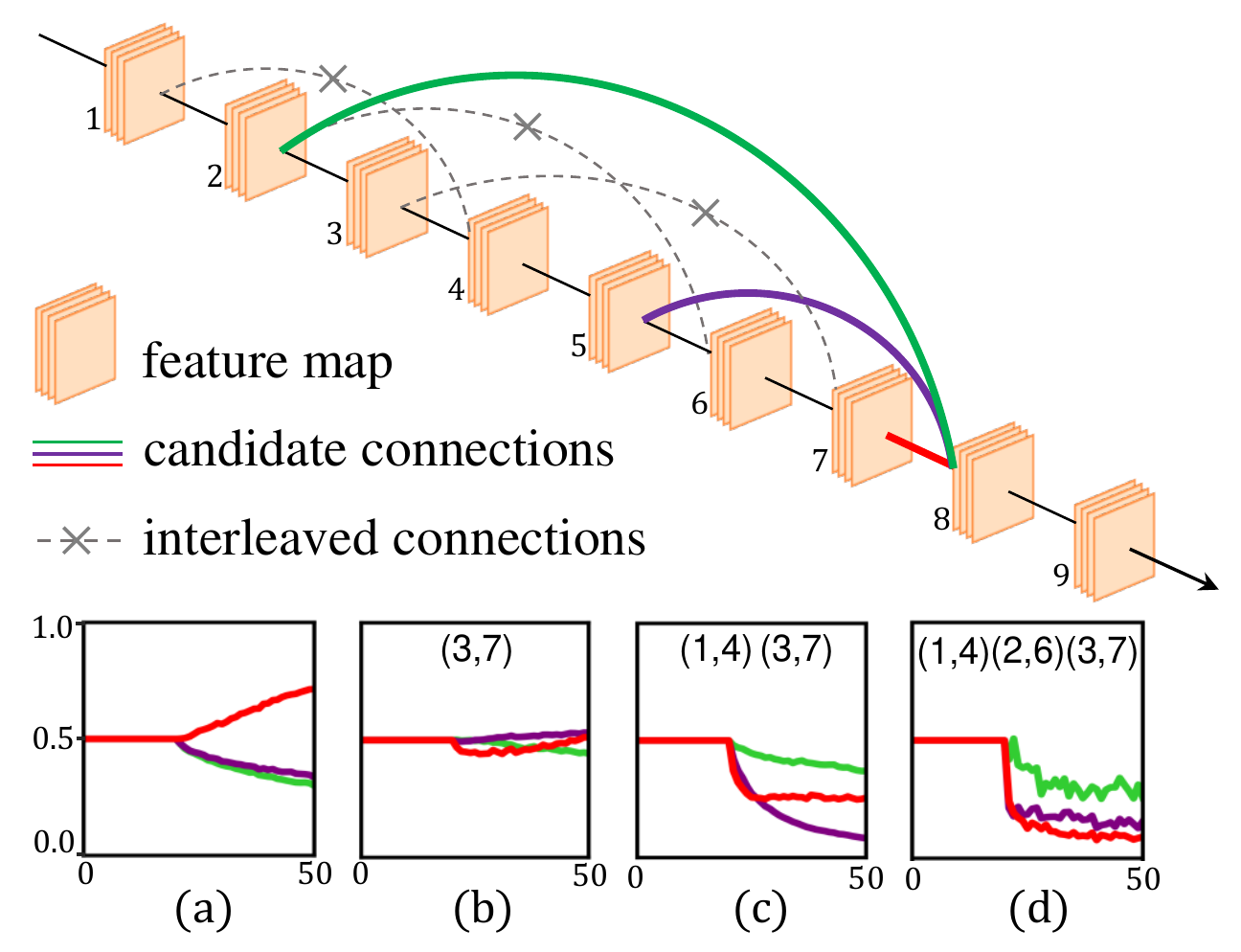}
\caption{Illustration of how interleaved connections contaminate the choice of neural network connections. The red, purple, and green links denote three candidate connections, among which the red one is the best choice (according to the individual evaluation). However, when interleaved connections (the dashed links) are added for the search sampling, the judgment of the candidates gradually becomes ridiculous. The bottom figures indicate the weight change of the three candidates throughout the search process (warm-up for 20 epochs), while $(a)$ is under the interleaving-free setting and $1$--$3$ interleaved connections are added for $(b)$--$(d)$. Specifically, the interleaved connection $(3,7)$ is added for $(b)$, the connections $(1,4)$, $(3,7)$ added for $(c)$, and $(1,4)$, $(2,6)$, $(3,7)$ added for $(d)$.}
\label{fig:introduction}
\end{figure}

Neural architecture search (NAS) is a research field that aims to automatically design deep neural networks~\cite{NASNet,AmoebaNet,metaQNN}. There are two important factors that define a NAS algorithm, namely, the search space that determines what kinds of architectures can appear, and the search strategy that explores the search space efficiently. Despite the rapid development of search algorithms which have become faster and more effective, the search space design is still in a preliminary status. In particular, for the most popular search spaces used in the community, either MobileNet-v3~\cite{mobilenetv3} or DARTS~\cite{liu2018darts}, the macro structure (\emph{i.e.}, how the network blocks are connected) is not allowed to change. Such a conservative strategy is good for search stability (\textit{e.g.}, one can guarantee to achieve good performance even with methods that are slightly above random search), but it reduces the flexibility of NAS, impeding the exploration of more complicated (and possibly more effective) neural architectures.

The goal of this paper is to break through the limitation of existing search spaces. For this purpose, we first note that the MobileNet-v3 and DARTS allow a cell to be connected to 1 and 2 precursors, respectively, resulting in relatively simple macro structures. In opposite, we propose a variant that each cell is connected to $L$ precursors ($L$ is 4, 6, or 8), and each connection can be either present or absent. We evaluate three differentiable NAS algorithms, namely DARTS~\cite{liu2018darts}, PC-DARTS~\cite{xu2020pcdarts}, and GOLD-NAS~\cite{GoldNAS} in the designed $L$-chain search space, and all of them run into degraded results. We perform diagnosis in the failure cases and the devil turns out to be the so-called \textbf{interleaved connections}, which refers to a pair of connections $(a,b)$ and $(c,d)$ that satisfies $a<c<b<d$. Figure~\ref{fig:introduction} shows an example that how interleaved connections affects the search results. With the increasing extent of interleaving, the search results gradually deteriorate, reflecting in the reduced accuracy and the weak operator gaining heavier weights. More examples are provided in the experiment.

The above observation motivates us to maximally eliminate the emerge of interleaved connections during the search procedure. This is easily done by performing interleaving-free sampling, where we group all candidate connections into $L$ groups and there exist no interleaved connections in every single group, based on which we periodically choose one group and perform regular NAS algorithms. This schedule can optimize the weight of every connection without suffering the issue of interleaved connections. Discretization and pruning are performed afterwards to derive the final architecture. The entire algorithm is named interleaving-free NAS, or \textbf{IF-NAS} for short.

We conduct experiments on ImageNet, a popular benchmark of NAS. In the newly proposed $L$-chain space, IF-NAS significantly outperforms three differentiable search baselines, DARTS, PC-DARTS and GOLD-NAS, and the advantage becomes more evident as the number of possible input blocks grows larger, \emph{i.e.}, heavier interleaving presents. 
Moreover, we evaluate IF-NAS in the existing search spaces of DARTS and GOLD-NAS, and show that it generalizes well to these easier search spaces.

In summary, the contributions of this paper are two-fold. \textbf{First}, we advocate for investigating the macro structure and put forward a novel search space for this purpose. The existing NAS methods cannot guarantee satisfying performance in this space. \textbf{Second}, we show that the major difficulty lies in dealing with the interleaved connections and hence propose an effective solution named IF-NAS. We hope that our efforts can inspire the NAS community to study the challenging new problem.

\section{Related Work}
 
\paragraph{Search Strategy}
 
Early NAS methods generally rely on individually evaluating the sampled sub-architectures under heuristic strategies, including reinforcement learning~\cite{NASNet} and the evolutionary algorithm~\cite{AmoebaNet}, which are computationally expensive. To accelerate the search, one-shot methods~\cite{Understanding, brock2017smash, SPOS} propose to represent the search space with a super-network, where the weights of all the candidate architectures are shared. Individual architecture training from scratch is avoided and the search cost is reduced by large magnitudes. Recently differentiable NAS (DNAS) has aroused great popularity in this field, which maps the discrete search space into a parameterized super network so that the search process can be executed by gradient descent. DARTS~\cite{liu2018darts}, as a pioneer differentiable framework, relaxes the search space by introducing updatable architecture parameters. The bi-level optimization is performed to update super-network weights and architectural parameters alternately. The target architecture is derived according to the distribution of architectural parameters. Due to its high efficiency, many works extend DNAS to more applications, including semantic segmentation~\cite{AutoDeeplab,zhang2019customizable}, object detection~\cite{fang2020fna++,guo2020hit}, \textit{etc}. Some DNAS works~\cite{cai2018proxylessnas,wu2019fbnet,DenseNAS} propose to integrate co-optimization with both accuracy and hardware properties into the search phase. In this paper, we target at promoting DNAS in terms of both flexibility and robustness. With the two factors improved, the final performance of DNAS can reach a higher level.

\paragraph{Eliminating Collapse in DNAS}
Though DNAS has achieved great success due to its high efficiency, many inherent problems exist with it and may cause collapse during search. A series of methods propose to improve DNAS from various perspectives. P-DARTS~\cite{PDARTS} bridges the gap between the super and searched network by gradually increasing the depths the networks. FairDARTS~\cite{chu2019fairdarts} improves the sampling strategy and breaks the two  indispensable factors of unfair advantages and exclusive competition. \cite{DAAS, GoldNAS} alleviate the discretization error by pushing the weights to sharp distributions. \cite{DARTS+, StacNAS, RobustDARTS} robustify the search process by introducing regularization techniques, including early stopping and observing eigenvalues of the validation loss Hessian, \textit{etc}. We reveal the degradation phenomenon in complicated search spaces when using the conventional DNAS method, and propose to suppress the interleaved connections during search. The proposed IF-NAS eliminates collapse most DNAS methods may encounter and shows superior performance.

\paragraph{Search Space}
Most existing NAS methods only explore in the micro search space, while the limited flexibility hinders further development of NAS. The cell-based space is firstly proposed in NASNet~\cite{NASNet}, which is widely adopted by the following works~\cite{ENAS, AmoebaNet, BayesNAS, liu2018darts}. This type of search space takes several nodes into one cell structure, which though eases the search procedure, suppresses many possible architectures with stronger feature extraction ability. \cite{liu2017hierarchical} proposes to search architectures under a hierarchical space which introduces flexible topology structures. Auto-DeepLab~\cite{AutoDeeplab} searches in a space with multiple paths and allows feature extraction in diverse resolutions. DenseNAS~\cite{DenseNAS} introduces densely connections in the search space and improves search freedom in terms of operators, depths and widths. GOLD-NAS~\cite{GoldNAS} liberates the restriction of the cell-based design and performs search in a global range. We further extends the search space complexity to explore architectures with more possibilities and potential. Though intractable is the search in such a complicated space, our proposed IF-NAS still shows evident effectiveness and advantages over other compared DNAS methods.

\section{Our Approach}

\subsection{Preliminaries: NAS in a Super-Network}

In neural architecture search (NAS), a deep neural network can be formulated into a mathematical function that receives an image $\mathbf{x}$ as the input and produces the desired information (\textit{e.g.}, a class label $y$) as the output. We denote the function to be $y=f(\mathbf{x};\boldsymbol{\alpha},\boldsymbol{\omega})$ where the form of $f(\cdot)$ is determined by a set of architectural parameters, $\boldsymbol{\alpha}$, and the learnable weights (\textit{e.g.}, the convolutional weights) are denoted by $\boldsymbol{\omega}$. The goal of NAS is to find the optimal architecture, $\boldsymbol{\alpha}^\star$, that leads to the best performance, \textit{i.e.},
\begin{eqnarray}
\label{eqn:goal}
{} & {\boldsymbol{\alpha}^\star}={\arg\min_{\boldsymbol{\alpha}}\mathbb{E}_{\left(\mathbf{x},y^\star\right)\in\mathcal{D}_\mathrm{val}}\!\left|y^\star-f(\mathbf{x})\right|}\\
\nonumber
\mathrm{s.t.} & {\boldsymbol{\omega}^\star(\boldsymbol{\alpha})}={\arg\min_{\boldsymbol{\omega}}\mathbb{E}_{\left(\mathbf{x},y^\star\right)\in\mathcal{D}_\mathrm{train}}\!\left|y^\star-f(\mathbf{x})\right|},
\end{eqnarray}
where $y^\star$ denotes the ground truth label.
Most often, $\boldsymbol{\alpha}$ takes discrete values, implying that solving Eqn~\eqref{eqn:goal} requires enumerating a large number of sampled architectures and performing individual evaluation. To accelerate, researchers propose to slack $\boldsymbol{\alpha}$ into a continuous form so that solving Eqn~\eqref{eqn:goal} involves optimizing a super-network, after which $\boldsymbol{\alpha}^\star$ is discretized into the optimal architecture for other applications.

In particular, this paper is built upon the differentiable search algorithms, in which the super-network is solved by computing the gradient with respect to $\boldsymbol{\alpha}$. We will introduce the details of optimization in the experimental part.

\subsection{Exploring a Complicated Search Space}

\begin{figure}
\centering
\includegraphics[width=.99\linewidth]{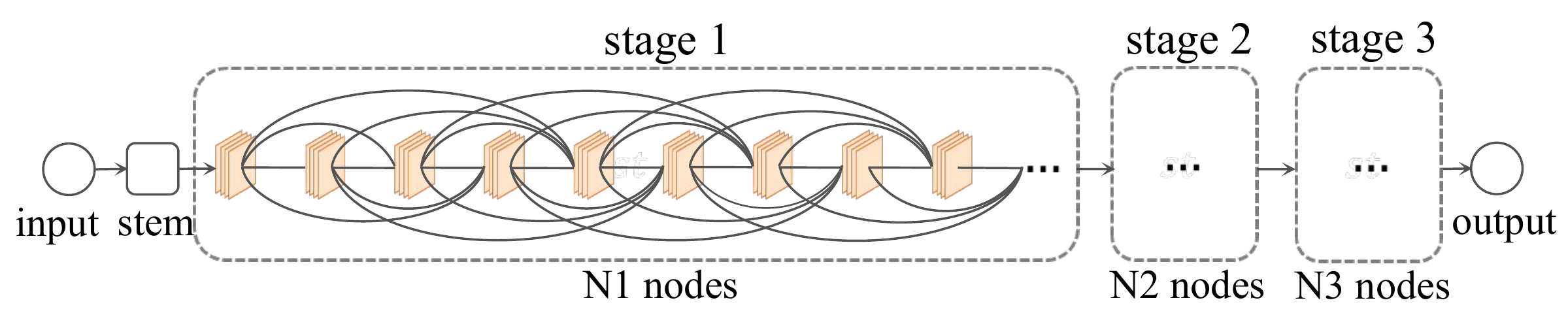}
\caption{The studied search space in this paper. Each layer (also called node) can be connected to $L$ precursors. 3 stages are to be searched with N1, N2 and N3 nodes respectively. For better visualization, we show the example of $L=4$, yet we also study $L=6$ and $L=8$ which are even more complicated.}
\label{fig:space}
\end{figure}

We design a search space shown in Figure~\ref{fig:space}. We define a fixed integer, $L$, indicating that each layer can be connected to $L$ precursors. For convenience, we name this space \textbf{$L$-chain space}. 
When $L=1$, it degrades to the chain-styled network (the backbone of MobileNet-v3). 
On the contrary, we study the cases of $L=4$, $L=6$, and even $L=8$, making the topology of the space much more complicated. We follow R-DARTS~\cite{RobustDARTS} to allow two candidate operations, \emph{i.e.} the $3\times3$ separable convolution and skip-connection. Throughout the remaining part of this paper, we use $\mathbb{S}^{(L)}$ to indicate the proposed space with $L$ precursor connections. For convenience, the connection between the $n$-th and $(n-l)$-th layers (nodes) is named the pre-$l$ connection of the $n$-th layer.

Before entering the algorithm part, we emphasize that we do not use additional rules to assist the architecture search, \textit{e.g.}, forcing all nodes to survive by preserving at least one connection. This raises new challenges to the search algorithm, because the depth of the searched architecture becomes quite indeterminate. Although some prior work has explored the option of optimizing the macro and micro architectures jointly~\cite{AutoDeeplab} or adding mid-range connections to the backbone~\cite{DenseNAS}, they still rely on the cell/block unit to perform partial search. Instead, we break the limitations of the unit, and allow search in a wider and macro space.

\begin{table}
\centering
\caption{Comparisons of the search space complexity.}
\label{tab.space_complexity}
\resizebox{0.47\textwidth}{!}{
\begin{tabular}{@{}cccccc@{}}
    \toprule
        \textbf{DARTS}
        & \textbf{GOLD-NAS} 
        & \textbf{$4$-chain}
        & \textbf{$6$-chain}
        & \textbf{$8$-chain}
      \\
    \midrule
      $1.1\times10^{18}$
      & $3.1\times10^{117}$
      & $1.8\times10^{116}$
      & $7.5\times10^{163}$
      & $1.6\times10^{204}$
      \\
    \bottomrule	
\end{tabular}
} 
\end{table}

In each edge, there are two candidate operators in our $L$-chain space. Each node receives at most $2L$ input feature maps. For the purpose of validity, each node has at least one input to be preserved. However, not every node may be used as an input node, and these nodes will be deleted, which means that the architectures are allowed to be very shallow. For a node $x_k (k>L)$, there are $2^{2L}-1$ input possibilities because each of the $2L$ operators can be on or off. As a result, there are $(2^2-1)\times(2^4-1)\times\cdot\cdot\cdot\times(2^{2(L-1)}-1)\times(2^{2L}-1)^{N-L}$ combinations if there are $N$ nodes in one stage. There are 3 stages to be searched in our space, Fig.~\ref{fig:space}, with the node number of 18, 20 and 18 respectively. Therefore, if $L$ is 4, 6 and 8,  there are about $1.8\times10^{116}$, $7.5\times10^{163}$ and $1.6\times10^{204}$ possible architectures respectively. The complexity comparison of popular spaces are shown in Tab.~\ref{tab.space_complexity}.

\begin{figure*}[t]
\begin{center}
\includegraphics[width=.90\linewidth]{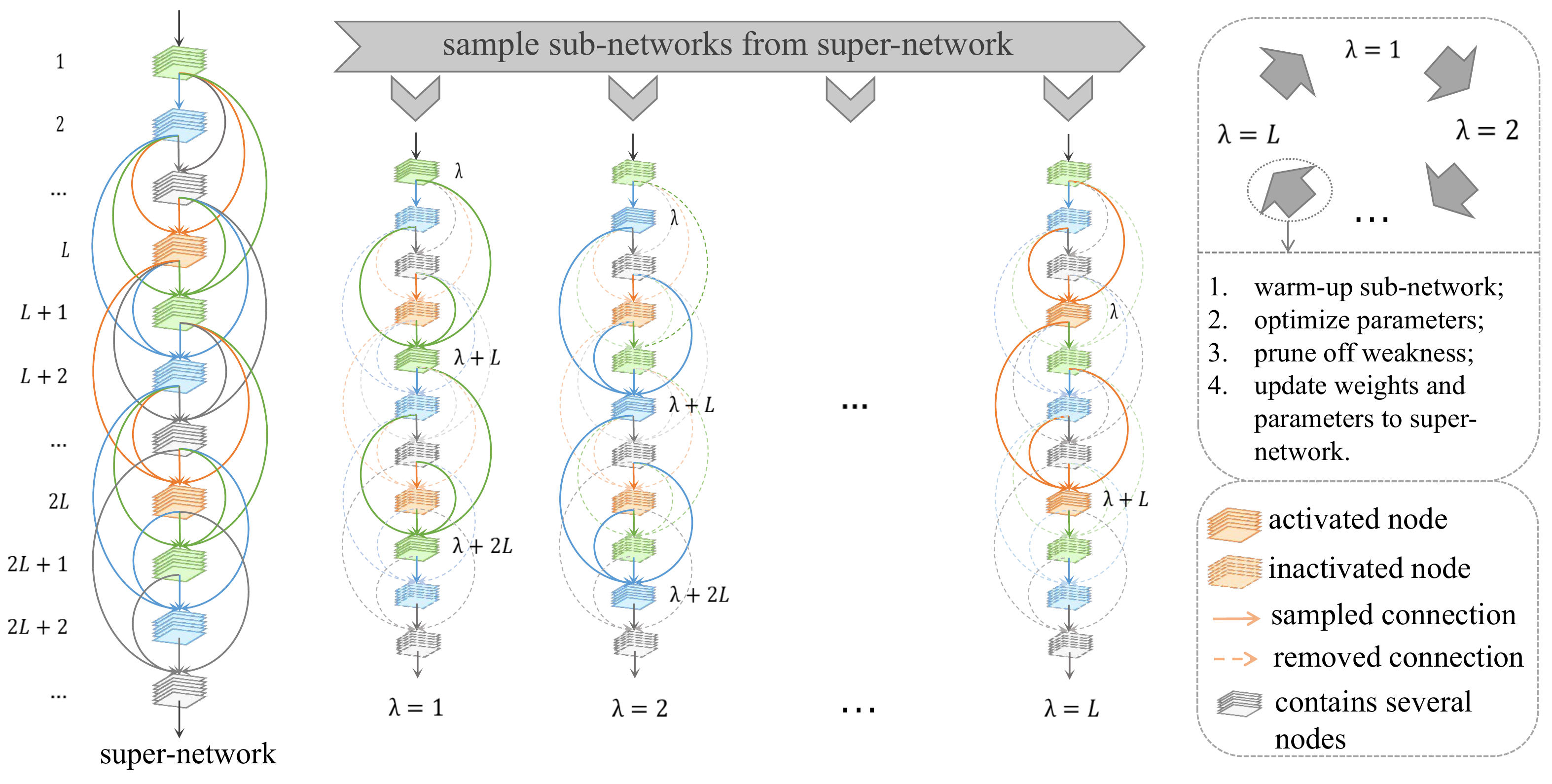}
\end{center}
\caption{The flowchart of IF-NAS. The key is to avoid interleaved connections in \textit{any} time. For this purpose, we repeat a loop with a length of $L$, and each time part of the connections remain active and all others are removed. The layers of the same color are always activated and inactivated together. This figure is best viewed in color.}
\label{fig:pipeline}
\end{figure*}

\subsection{Failure Cases and Interleaved Connections}

We first evaluate DARTS~\cite{liu2018darts} and GOLD-NAS~\cite{GoldNAS} in the search space of $\mathbb{S}^{(4)}$. 
On the ImageNet-1k dataset, trained for 250 epochs for each network, DARTS and GOLD-NAS report top-1 accuracy of $74.7\%$ and $75.3\%$, respectively, and the FLOPs of both networks are close to $600\mathrm{M}$ (\textit{i.e.}, the mobile setting), Tab.~\ref{tab.main_results}. 
In comparison, the simple chain-styled architecture (each node is only connected to its direct precursor) achieves $74.9\%$ with merely $520\mathrm{M}$ FLOPs.
More interestingly, if we only preserve one input for each node, DARTS and GOLD-NAS report completely failed results of $70.2\%$ and $71.2\%$ (trained for 100 epochs), which are even much worse than preserve one input randomly, Tab.~\ref{tab.ablation_keep1}.

We investigate the searched architectures, and find that both DARTS and GOLD-NAS tend to find long-distance connections. This decreases the depth of the searched architectures, however, empirically, deeper networks often lead to better performance. 
This drives us to rethink the reason that the algorithm gets confused. For this purpose, we randomly choose an intermediate layer from the super-network and the algorithm needs to determine the preference among its precursors.

We start from the simplest situation that all network connections are frozen (the architectural weights are fixed) besides the candidate connections. We show the trend of three (pre-$1$, pre-$3$, pre-$6$) connections in Figure~\ref{fig:introduction}. One can observe that the pre-$1$ connection overwhelms other two connections, aligning with our expectation that a deeper network performs better. However, when we insert one connection that lies between these candidates, we observe that the advantage of the pre-$1$ connection largely shrinks, and a long training procedure is required for it to take the lead. The situation continues deteriorating if we insert more connections into this region. When two or more connections are added, the algorithm cannot guarantee to promote the pre-$1$ connection and, sometimes, the ranking of the three connections is totally reversed. 

The above results inspire us that two connections are easily interfered by each other if the covering regions (\textit{i.e.}, the interval between both ends) overlap. Hereafter, we name such pair of connections \textbf{interleaved connections}. Mathematically, denote a connection that links the a-th and b-th nodes as $(a, b)$, where $b-a\ge 2$. Overlook the candidate connections of the activated nodes, two connections, $(a,b)$ and $(a',b')$, interleave if and only if there exists at least one integer $d$ that satisfies $a<d+1/2<b$ and $a'<d+1/2<b'$.
Intuitively, for a real number $x$, if $\{x| a < x < b\} \cap \{x | a' < x < b'\} \neq \emptyset$ is satisfied, the interleaved connection occurs.


As a side comment, the DARTS space is also impacted by the interleaved connections, which partly cause the degradation problem observed in prior work~\cite{DARTS+,PDARTS,Cellbased}. However, since the search space is cell-based, the degradation does not cause dramatic accuracy drop. In the experiments, we show that our solution, elaborated in the next part, generalizes well to the DARTS space.

\subsection{Interleaving-Free Sampling}

Following the above analysis, the key to design the search algorithm is to avoid interleaved connections, meanwhile ensuring that all connections can be considered. For this purpose, we propose IF-NAS, a sampling-based approach that each time optimizes a interleaving-free sub-super-network from the super-network.

This is implemented by partitioning the layers into $L$ groups, $\mathcal{G}_1,\mathcal{G}_2,\ldots,\mathcal{G}_L$, according to their indices modulo $L$. In other words, the distance between any consecutive layers that belong to the same group is exactly $L$. When any group $\mathcal{G}_{\lambda}$ is chosen, we obtain an interleaving-free sub-super-network by preserving (i) the main backbone (all pre-$1$ connections) and (ii) all connections that end at any layer in $\mathcal{G}_{\lambda}$. We denote the sub-super-network by $\mathbb{S}_{\lambda}^{(L)}$ for $\lambda=1,2,\ldots,L$. Correspondingly, the optimization goal can be written as:
\begin{equation}
\label{eqn:loss1}
\min_{\boldsymbol{\alpha},\boldsymbol{\omega}}\mathbb{E}_{\left(\mathbf{x},y^\star\right)\in\mathcal{D}}\left|y^\star-f\!\left(\mathbf{x}\mid\mathbb{S}_{\lambda}^{(L)}\right)\right|.
\end{equation}

A straightforward search procedure repeats the loop of $L$ sub-super-networks, as shown in Figure~\ref{fig:pipeline}. Training each sub-super-network is an approximation of optimizing the entire super-network, $\mathbb{S}^{(L)}$, but, since the sub-super-network are fixed, the co-occurring connections may gain unexpected advantages over other ones. To avoid it, we add a warm-up step to the beginning of each loop, in which the entire super-network is trained (all connections are activated, but the architectural parameters are not updated). Throughout the remainder of the loop, we train the $L$ sub-super-networks orderly. We set the length of each step to be $100$ iterations (\textit{i.e.}, mini-batches). Note that this number shall not be too large, otherwise the gap between the sampled and non-sampled connections will increase and harm the search performance.

\subsection{Discretization and Pruning}

The last step is to determine the final architecture. In a complicated search space, this is not simple as it seems, because many connections may have moderate weights (\emph{i.e.}, not close to $0$ or $1$). For such a connection, either pruning it or promoting it can bring significant perturbation on the super-network. Therefore, we follow the idea of prior works~\cite{chu2019fairdarts,GoldNAS,DAAS} to first perform a discretization procedure to push the weights towards either $0$ or $1$, after which pruning is much safer.
We introduce two kinds of architectural parameters, the connectivity of edges are determined by $\boldsymbol{\beta}$ and operators are determined by $\boldsymbol{\alpha}$.

The key of discretization is to add a regularization term to Eqn. (\ref{eqn:loss1}). The term penalizes the moderate weights of edges and operators, represented by $\boldsymbol{\beta}$ and $\boldsymbol{\alpha}$ respectively. Once these weights are less than $0.01$, the corresponding edges or operators will be pruned off. The regularization term is computed by:

\begin{footnotesize}
\begin{equation}
\begin{split}
    \mathcal{R}(\alpha,\beta)&=\mu_1\cdot \sum_{j\le N}\sum_{0\le j-L\le i<j}\ln(1+g(\beta_{i,j})/\overline{g(\beta_{i,j})})\\
    &+\mu_2\cdot \sum_{j\le N}\sum_{0\le j-L\le i<j} \sum_{o\in \mathcal{O}}\ln(1+g(\alpha^o_{i,j}|\beta_{i,j})/\overline{g(\alpha^o_{i,j}|\beta_{i,j})}),
\end{split}
\end{equation}
\end{footnotesize}
where $N$ is the number of layers, and $g(\cdot)$ is the activate function $sigmoid$. 
$g(\alpha|\beta)$ means to only include operators on $\boldsymbol{\beta}$ that have not been pruned off. $\overline{g(\cdot)}$ is the average of $g(\cdot)$.
$\mathcal{R}(\alpha,\beta)$ is multiplied by a factor of $\mu$ and added to the cross-entropy loss, so that optimizing the overall loss will not only improve the recognition accuracy of the super-network, but also push the weights of all connections towards $0$ or $1$.

When the weight of a connection is sufficiently small, we prune it permanently from the super-network. This merely impacts the super-network itself, but the computation of $\overline{g(\cdot)}$ changes and will push the next weak operator to $0$. This process iterates until the complexity (\textit{e.g.}, FLOPs) of the super-network achieves the lower-bound.




\section{Experiments}
In this section, we first introduce experimental details and implementation details. Next, we introduce the results and analysis of IF-NAS and other advanced methods in our $L$-chain space under different settings.
Finally, we compare $L$-chain spaces with 2 cell-based micro spaces using IF-NAS and other 2 search strategies. 

\subsection{Implementation Details}
\textbf{The Dataset.}
We use the large-scale ImageNet dataset (ILSVRC2012) to evaluate the models. ImageNet contains 1,000 object categories, which consists of 1.3M training images and 50K validation images. The images are almost equally distributed over all classes. Unless specified, we apply the mobile setting, in which the input image is set to $224\times224$ and the number of multiply-add operations (MAdds) does not exceed 600M. We randomly sample 100 classes from the original ImageNet dataset to perform studying and analysis experiments, ImageNet-100 for short.


\textbf{The Search Settings.}
Before searching, IF-NAS warm-ups the super-network for 20 epochs, with only the super-network weights updated and the architectural parameters frozen. Then we start to update $\beta$ firstly, and after 10 epochs, start to update $\alpha$. Super-network weights and architectural parameters are both optimized by the SGD optimizer. 
We gradually prune off weak edges and operators with threshold of $0.01$ until the MAdds of the retained architecture meets the mobile setting, \emph{i.e.}, 600M. 

\textbf{The Training Settings.}
We evaluate the searched architectures following the setting of PC-DARTS~\cite{xu2020pcdarts}. Each searched architecture is trained from scratch with a batch size of 1024 on 8 Tesla V100 GPUs. A SGD optimizer is used with an initial learning rate of 0.5, weight decay ratio of $3\times10^{-5}$, and a momentum of 0.9. Other common techniques including label smoothing, auxiliary loss, and learning rate warm-up for the first 5 epochs are also applied.

\subsection{Results on $L$-chain Search Space}

\begin{table}
\fontsize{8.5}{10.0}\selectfont
\renewcommand\arraystretch{1.01}
\centering
\caption{Results of popular methods on our enlarged complicated space: comparison of classification test error (\%) trained on ImageNet-1k for 250 epochs under the mobile setting.}
\label{tab.main_results}
\resizebox{0.48\textwidth}{!}{
\begin{tabular}{@{}lcccccc@{}}
    \toprule
        \multirow{2}{*}{\textbf{Setting}}
        & \multirow{2}{*}{\textbf{Method}}
        & \multicolumn{2}{c}{\textbf{Test Err. (\%)}}
        & \textbf{Params}
        & $\times+$
      \\
        \cmidrule(lr){3-4}
        &
        & \textbf{top-1}
        & \textbf{top-5}
        & \textbf{(M)}
        & \textbf{(M)}
        \\
    \midrule
        \multirow{4}{*}{\textbf{$L$ = 4}}
        &DARTS~\cite{liu2018darts}  &25.3  &8.1  &5.1  &589 
        \\
        &PC-DARTS~\cite{xu2020pcdarts} &24.7  &7.5  &5.3  &593  
        \\
      & GOLD-NAS~\cite{GoldNAS}  &24.7  &7.5  &5.5  &591 
        \\
      & IF-NAS &\textbf{24.3} & \bf 7.4  &5.3  &592
        \\
    \midrule		
     \multirow{4}{*}{\textbf{$L$ = 6}}
        &DARTS~\cite{liu2018darts}  &25.7  &8.1  &5.1  &587 
        \\
        &PC-DARTS~\cite{xu2020pcdarts} &24.9  &7.7  &5.2  &586 
        \\
      & GOLD-NAS~\cite{GoldNAS} &25.0  &7.8  &5.4  &596 
        \\
      & IF-NAS &\textbf{24.4} & \bf7.4  & 5.2  &598  
        \\
    \midrule		
     \multirow{4}{*}{\textbf{$L$ = 8}}
        &DARTS~\cite{liu2018darts}  &25.9  &8.2  &5.2  &591 
        \\
        &PC-DARTS~\cite{xu2020pcdarts} &25.1  &7.9  &5.2  &582 
        \\
      & GOLD-NAS~\cite{GoldNAS} &25.1  &7.9 &5.3  &592 
        \\
      & IF-NAS &\textbf{24.3} & \bf7.3  & 5.4  &594  
        \\
    \bottomrule
\end{tabular}
}
\end{table}

We study our method IF-NAS on the proposed macro $L$-chain search space, and compare it with three other popular DNAS frameworks, including DARTS~\cite{liu2018darts}, PC-DARTS~\cite{xu2020pcdarts} and GOLD-NAS~\cite{GoldNAS}. 
For a fair comparison, the hyper-parameters are kept the same for all the studied methods. 
Without pruning gradually, DARTS and PC-DARTS derive the final architectures by removing weak edges and operators after searching until they meet the requirements of mobile setting. All the searched architectures are re-trained from scratch on ImageNet-1k for 250 epochs.

We perform three sets of experiments by setting $L$, which indicates the precursors number of each layer can be connected, to 4, 6 and 8 respectively, and show the results in Tab.~\ref{tab.main_results}.
When $L$ is 4, DARTS gets the worst performance as expected. Compared with DARTS, PC-DARTS and GOLD-NAS get better results of $24.7\%$ and $24.7\%$ with similar MAdds. Our IF-NAS gets the best result of $24.3\%$ with equivalent Params and MAdds. This proves that IF-NAS can work better in such a enlarged complicated space compared to the methods. 

When $L$ is 6, DARTS achieves the result of $25.7\%$. PC-DARTS and GOLD-NAS get results of $24.9\%$ and $25.0\%$ respectively. And our IF-NAS gets the best result of $24.4\%$ with similar Params and MAdds. Note that the results of the compared 3 frameworks get worse when $L$ increases to 6 compared to $L$ as 4. However, IF-NAS nearly maintains the performance. When $L$ increases, more interleaved connections occur, and bring more interference to the search process. The compared 3 methods, which do not suppress interleaved connections during search, are more severely effected and achieve worse results. However, IF-NAS samples interleaving-free sub-networks and updates the architectural parameters more accurately. This further confirms that our IF-NAS has great potential to handle complicated spaces. 

When set $L$ to 8, IF-NAS still achieves a promising result of $24.3\%$, which is comparable to the former two settings. The compared ones get similar or weaker results. 
Due to the increase in complexity, the performances of PC-DARTS and GOLD-NAS continue to decrease. But the MAdds of these architectures are close to 600M, which prevents performance from declining.
\begin{table}
\fontsize{8.5}{10.5}\selectfont
\centering
\caption{Results of reserving one input for each node. all architectures are trained on ImageNet-1k for 100 epochs.}
\label{tab.ablation_keep1}
\resizebox{0.48\textwidth}{!}{
\begin{tabular}{@{}lcccccc@{}}
    \toprule
        \textbf{Setting}
        & \textbf{random}
        & \textbf{DARTS}
        & \textbf{GOLD-NAS}
        & \textbf{IF-NAS}
      \\
    \midrule
      \textbf{$L$ = 4}
      & $27.6\pm0.3$
      & 29.8
      & 28.8
      & 26.7
      \\
      \textbf{$L$ = 6}
      & $28.0\pm0.3$
      & 30.2
      & 29.0
      & 26.9
      \\
     \textbf{$L$ = 8}
      & $28.9\pm0.4$
      & 31.6
      & 30.8
      & 27.3
      \\
    \bottomrule	
\end{tabular}
}
\end{table}

If we do not derive architectures with MAdds and preserve one input for each node, the results are shown in Tab.~\ref{tab.ablation_keep1}. Interestingly, without the protection of keeping so many MAdds, DARTS and GOLD-NAS show completely failed results. When $L$=4, the results of DARTS and GOLD-NAS are $29.8\%$ and $28.8\%$, which are even much worse than the randomly generated architectures. When $L$ increases to 8, the performances of DARTS and GOLD-NAS further degrade. The architectures searched without suppressing interleaved connections only preserve very few nodes, which damages the final performance dramatically. On the contrary, IF-NAS still achieves comparable results even without any MAdss restriction.

The above experiments imply interleaving connections cause non-negligible impacts to DNAS methods when the search space grows larger and more complicated. IF-NAS is able to handle this challenge by the proposed interleaving-free sampling.


\begin{figure}[thbp]
    \centering
    \begin{minipage}{0.49\textwidth}
        \centering
        \includegraphics[width=0.99\textwidth]{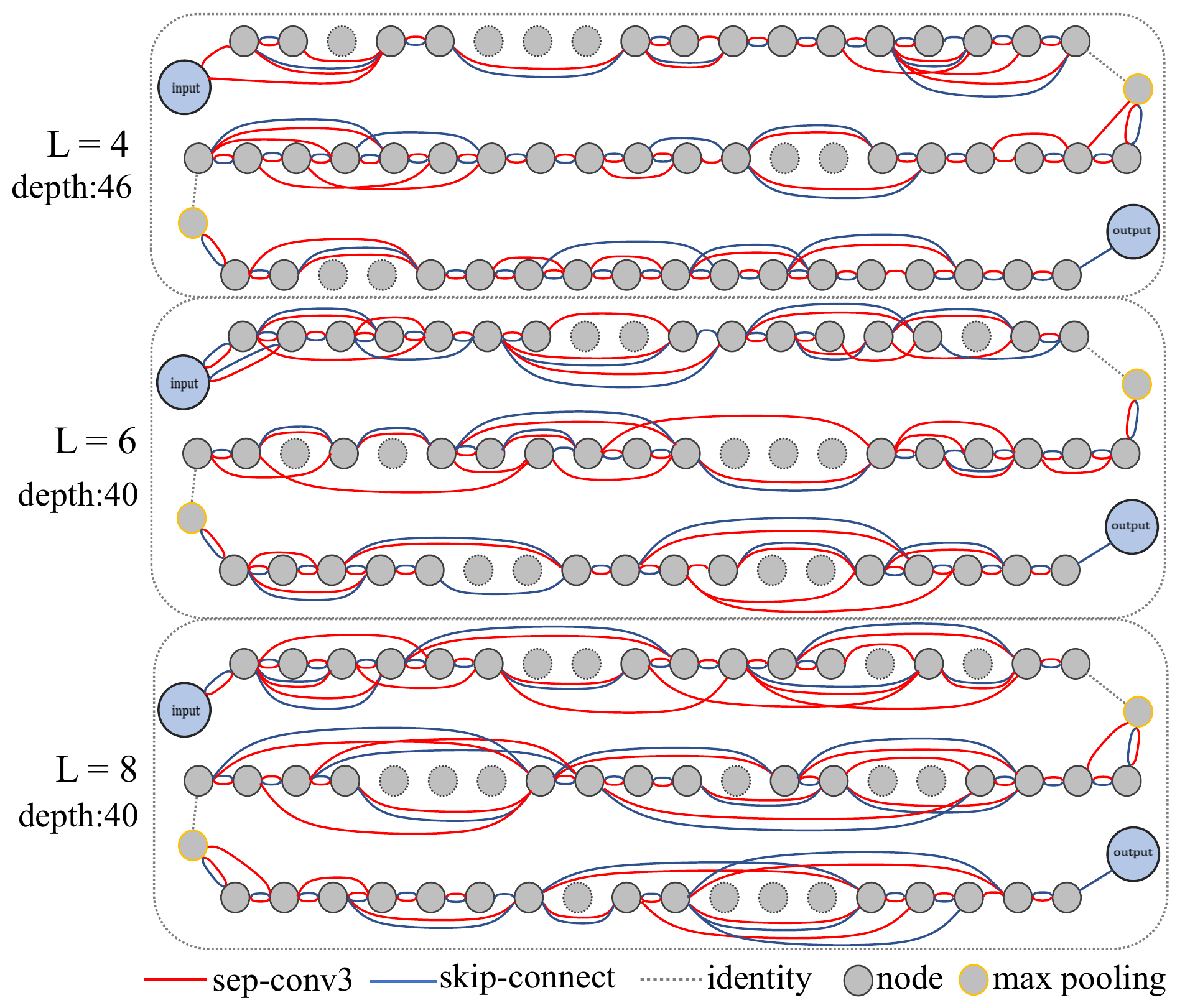}
        \captionof{figure}{Architectures searched with IF. The reserved connections are mainly serial pattern, which guarantees the depth of architectures.}
        \label{fig:architecture_withIF}
    \end{minipage}%
    \hspace{1pt}
    \begin{minipage}{0.49\textwidth}
        \centering
        \includegraphics[width=0.95\textwidth]{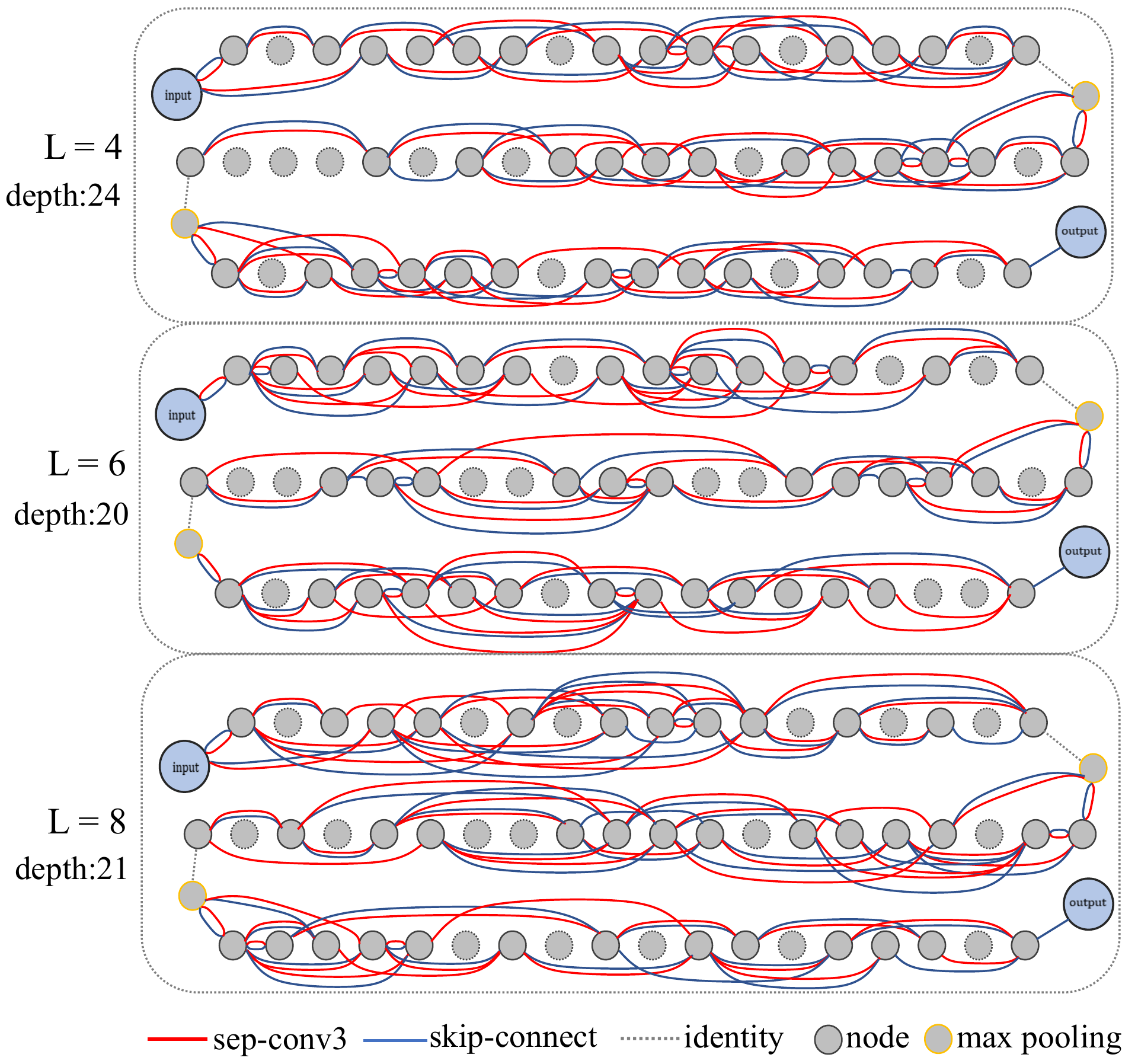}
        \captionof{figure}{Architectures searched without IF. The reserved connections are mainly parallel pattern, resulting in a short longest-path at each stage, that is, the depth of networks are shallow.}
        \label{fig:architecture_withoutIF}
    \end{minipage}%
\end{figure}

\begin{table}
\centering
\caption{Results of exploring the effectiveness of IF sampling on the $L$-chain space. All the architectures are trained on ImageNet-1k for 100 epochs.}
\label{tab.ablation_if}
\resizebox{0.42\textwidth}{!}{
\begin{tabular}{@{}lcccccc@{}}
    \toprule
        \textbf{Setting}
        & \textbf{Depth}
        & \textbf{Params (M)} 
        & \textbf{Err. (\%)}
      \\
    \midrule
      w/o IF
      & $22.5\pm 2.0$
      & $5.2\pm 0.2$
      & $26.7\pm 0.3$
      \\
      w/ IF
      & $43\pm 3$
      & $ 5.3 \pm 0.1$
      & $26.0\pm 0.3$
      \\
    \bottomrule	
\end{tabular}
}
\end{table}

\subsection{IF Sampling Effectiveness on $L$-chain Space}

In this part, we design a comparative experiment to verify the effectiveness of IF sampling. 
We search for 4 times independently for each setting, and the results are shown in Tab.~\ref{tab.ablation_if}. The \textbf{Depth} in Tab.~\ref{tab.ablation_if} refers to the depth summation of 3 stages, which represents the longest path length of the network. And all architectures are train on ImageNet-1k for 100 epochs.

\begin{figure}
\centering
\includegraphics[width=8.5cm]{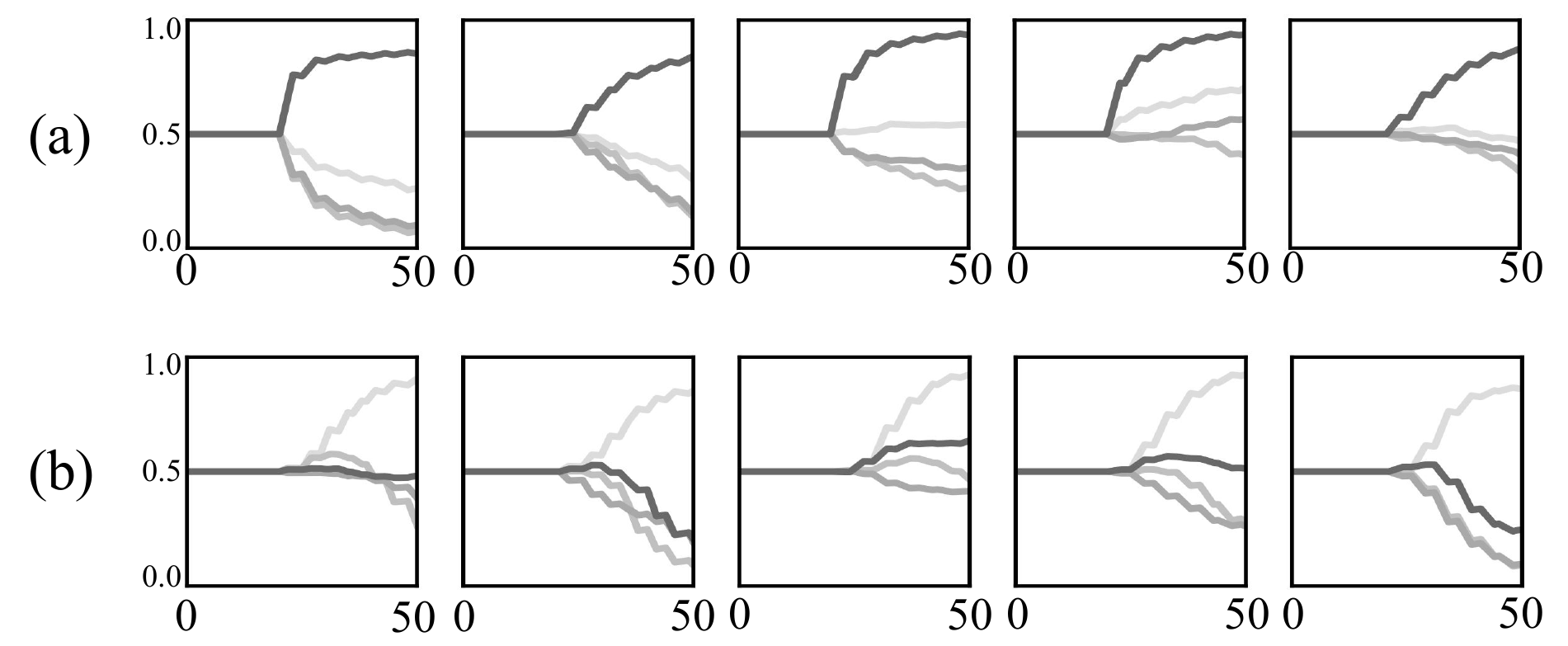}
\caption{(a): The weight changing curve of candidate inputs with IF. (b): The weight changing curve of candidate inputs without IF. The darker the color, the closer the corresponding candidate inputs are to the activated node.}
\label{fig:weight_change}
\end{figure}

We find the accuracy of the architectures searched without IF (``w/o IF'') sampling is much worse than that with IF (``w/ IF'') sampling by a margin of $0.7\%$ with similar Params. It is worth noting that the depth of w/o IF is much smaller than w/ IF, because the interleaved connections produce serious interference to the search process. This verifies that our IF-NAS can effectively handle interleaved connections and guarantee the search performance.


The visualizations of searched architectures are shown in Fig.~\ref{fig:architecture_withIF} and Fig.~\ref{fig:architecture_withoutIF}. We can find that w/ IF preserves many serial pattern connections and w/o IF  mainly preserves parallel connections, as a result, the architectures of w/ IF is much deeper than w/o IF. Fig.~\ref{fig:weight_change} shows some weight change curves of w/ IF and w/o IF. One can see that w/ IF can select more closer inputs, which derives deeper architectures. w/o IF prefers farther nodes, which usually leads to parallel connections and discards more nodes.

\subsection{Effects of Interleaved Connections}

\begin{table}
\fontsize{8.0}{10.5}\selectfont
\centering
\caption{Results of retaining architectures searched with different number of interleaved connections (IC) involved.}
\label{tab.ablation_twisting}
\resizebox{0.45\textwidth}{!}{
\begin{tabular}{@{}ccccc@{}}
    \toprule
        \textbf{IC Number}
        & \textbf{Depth}
        & \textbf{Params (M)}
        & \textbf{Err. (\%)}
      \\
    \midrule
      0
      & $41.5\pm 2.5$
      & $5.0 \pm 0.1$
      & $19.1\pm 0.2$
      \\
      1
      & $26.0\pm 3.0$
      & $4.9\pm0.3$
      & $19.7\pm0.3$
      \\
       2
      & $24.5\pm 2.5$
      & $5.0 \pm 0.3$
      & $19.9\pm 0.4$
      \\
       3
      & $24.0\pm 3.0$
      & $5.0\pm 0.2$
      & $20.0\pm 0.3$
      \\
    \bottomrule	
\end{tabular}
} 
\end{table}

We allow various numbers (1/2/3) of interleaved connections to be sampled during search to understand the effects they may bring.
The searched architectures are trained on ImageNet-100 for 100 epochs and the results are shown in Tab.~\ref{tab.ablation_twisting}. We find with more interleaved connections added, the performance degrades more. This shows more interleaved connections cause greater impact on the search. It implies that Interleaving-Free is necessary in such a flexible space. Interestingly, the performance drops greatly even through one interleaved connection is sampled during searching, which demonstrates that as long as the interleaved connection is involved, interference is generated.


\subsection{Comparison with Cell-based Micro Spaces}


\begin{figure}
\centering
\includegraphics[width=8.0cm]{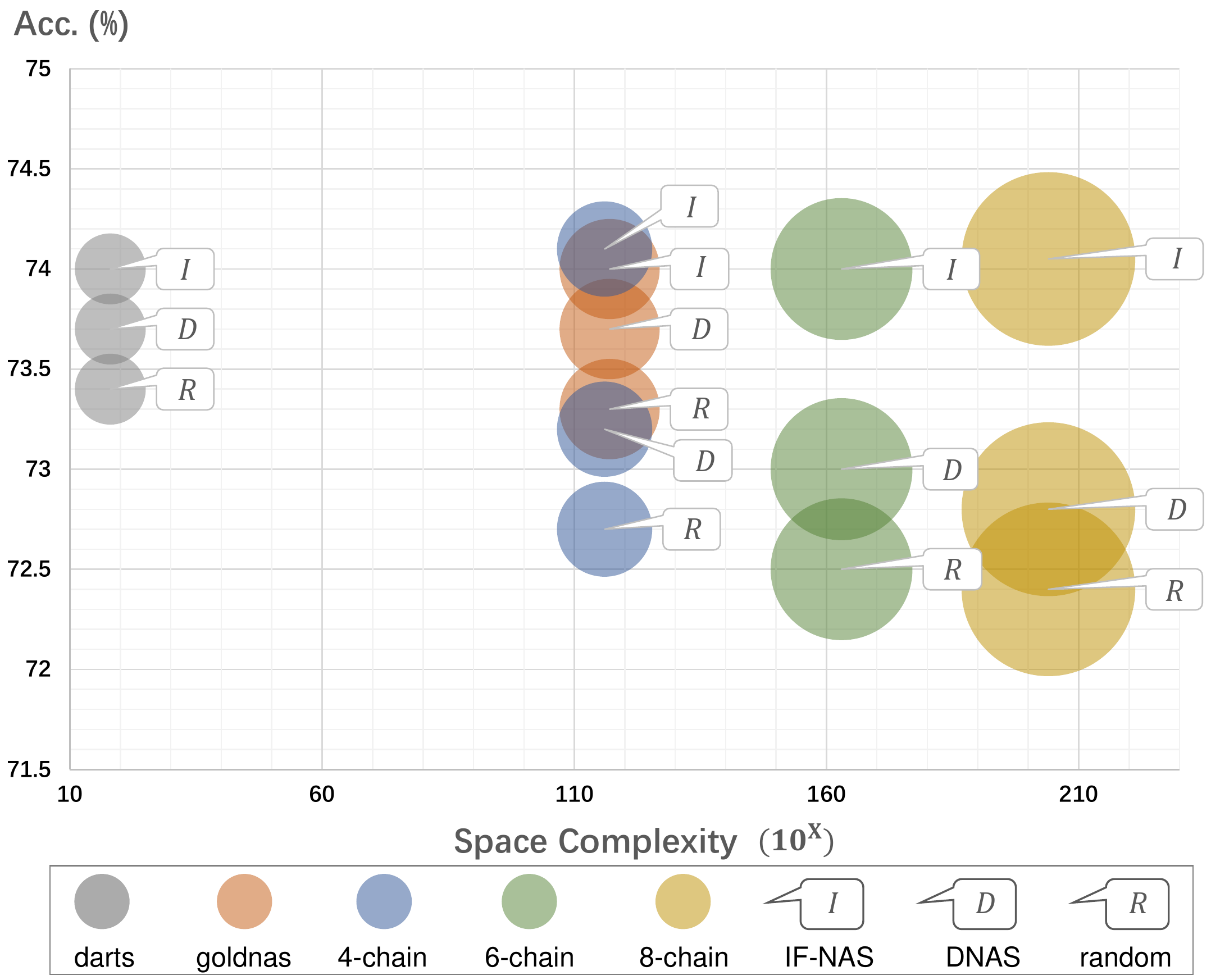}
\caption{Results on different spaces using 3 search strategies. DNAS indicates the differential-based search without IF. The random indicates random search.}
\label{fig:acc_complex}
\end{figure}

In this section, we test IF-NAS in two micro search spaces, DARTS space referred to as $space_D$ and GOLD-NAS space referred to as $space_G$. The two spaces adopt the same $3\times3$ separable convolution and skip-connection operators as ours, thus that the comparison is fair. The results are shown in Fig.~\ref{fig:acc_complex}. The horizontal axis denotes the space complexity (the exponential base of 10) and the vertical axis denotes the accuracy trained on ImageNet-1k for 100 epochs. The bubble sizes imply the complexity of spaces. The results on $space_D$ and $space_G$ are close among 3 strategies (even with random search), which implies the 2 spaces are inflexible and have a protection mechanism (the cell design) to help the search methods. The $L$-chain spaces are more flexible and difficult so that the gaps between IF-NAS and others in $L$-chain spaces are much larger. Moreover, the larger the spaces, the worse the performances of DNAS and random search. However, our IF-NAS provides all best results in the 5 spaces, which shows that IF-NAS are robust and generalize well to micro spaces.

We also test our IFNAS by training 250 epochs under the mobile setting and the results are shown in Table~\ref{tab:micro_results}. IFNAS achieves $24.2\%$ test error in $space_D$, which is same as PCDARTS (a specially designed method in this space). In $space_G$, our IFNAS-G1 beats GOLD-I by $0.2\%$ with $0.2$M fewer parameters and our IFNAS-G2 surpasses GOLD-X with $0.4$M fewer parameters.

\begin{table}[]
\fontsize{8.5}{10.5}\selectfont
\centering
\setlength{\tabcolsep}{1.0mm}
    \caption{Results on two micro search spaces.}
    \begin{tabular}{ccc|ccc}
    \toprule
         \multicolumn{3}{c|}{\textbf{$space_D$}}  &\multicolumn{3}{c}{\textbf{$space_G$}} \\ \hline
         arch.  & para. (M)  & Err. (\%)  &arch.  & para. (M)  &Err. (\%) \\ \hline
         DARTS  &4.7    &   26.7   & GOLD-I   & 5.3   &  24.7  \\
         PDARTS   &4.9   & 24.4    & GOLD-X    & 6.4   &24.3\\
         PCDARTS   &5.3   &24.2     & IF-NAS-G1  & 5.1   &24.5   \\
         IF-NAS    &5.3   &\textbf{24.2}     &IF-NAS-G2   &6.0   &\textbf{24.2}  \\
    \bottomrule 
    \end{tabular}
    
    \label{tab:micro_results}
\end{table}

\subsection{More Examples of Interleaved Connections}

\begin{figure}[]
    \centering
    \includegraphics[width=0.8\linewidth]{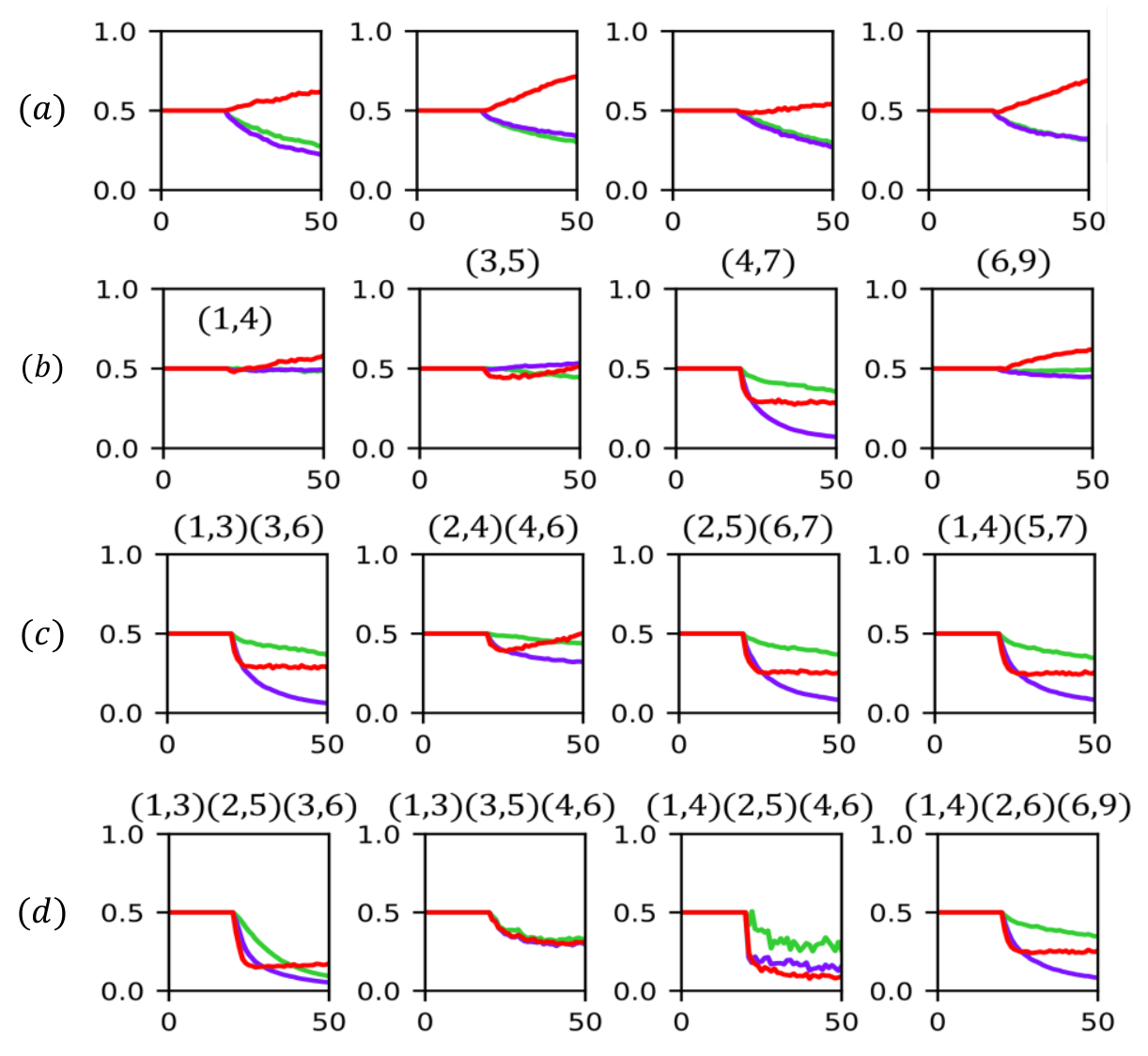}
    \caption{More examples of adding different numbers of interleaved connections to impact the NAS algorithm. (a) is under the interleaving-free setting and 1-3 interleaved connections are added for (b)-(d).}
    \label{fig:noise_all}
\end{figure}

In Section 3.3, we show failure cases from the simplest situation. Here we perform more experiments to show the influence of interleaved connections and the weight change curves are shown in Figure~\ref{fig:noise_all}. In Fig.~\ref{fig:noise_all} (a), we run 4 times independently under interleaving-free setting and the nearest candidate inputs have a dominant weight at this time. In Fig.~\ref{fig:noise_all} (b)-(d), We run 4 times independently and each time we add 1-3 different interference connections. One can see that the more interleaved connections are added, the harder it it the nearest candidate input is picked.

\section{Conclusions}

This work extends the design of search space towards higher flexibility by adding lots of long-range connections to the super-network. The long-range connections raise new challenges to the search algorithm, and we locate the problem to be the interference introduced by interleaved connections during search. We propose IF-NAS based on this observation, which makes a schedule to sample sub-networks and thus guarantees that interleaved connections do not occur.

Searching in a more complicated space has been of critical importance for NAS. Our research delivers the message that new properties/challenges of NAS emerge when the search space is augmented. We advocate for more efforts in this direction to improve the ability of the search algorithms.

\vspace{0.1cm}\noindent
\textbf{Limitations of this work.} First, the architectures searched in the complicated spaces do not show advantages than the existing cell-based ones, which implies that the greater flexibility we introduced does not necessarily lead to better results but greater difficulties.  Second, IF-NAS needs an extra warm up process to make up for the accuracy drop of super-network caused by sampling operation.

{\small
\bibliographystyle{ieee_fullname}
\bibliography{egbib}

\begin{thebibliography}{10}\itemsep=-1pt

\bibitem{metaQNN}
Bowen Baker, Otkrist Gupta, Nikhil Naik, and Ramesh Raskar.
\newblock Designing neural network architectures using reinforcement learning.
\newblock In {\em ICLR}, 2017.

\bibitem{Understanding}
Gabriel Bender, Pieter{-}Jan Kindermans, Barret Zoph, Vijay Vasudevan, and
  Quoc~V. Le.
\newblock Understanding and simplifying one-shot architecture search.
\newblock In {\em ICML}, 2018.

\bibitem{GoldNAS}
Kaifeng Bi, Lingxi Xie, Xin Chen, Longhui Wei, and Qi Tian.
\newblock {GOLD-NAS:} gradual, one-level, differentiable.
\newblock {\em CoRR}, abs/2007.03331, 2020.

\bibitem{brock2017smash}
Andrew Brock, Theodore Lim, James~M Ritchie, and Nick Weston.
\newblock {SMASH}: one-shot model architecture search through hypernetworks.
\newblock In {\em ICLR}, 2018.

\bibitem{cai2018proxylessnas}
Han Cai, Ligeng Zhu, and Song Han.
\newblock {ProxylessNAS}: Direct neural architecture search on target task and
  hardware.
\newblock In {\em ICLR}, 2019.

\bibitem{PDARTS}
Xin Chen, Lingxi Xie, Jun Wu, and Qi Tian.
\newblock Progressive differentiable architecture search: Bridging the depth
  gap between search and evaluation.
\newblock In {\em ICCV}, 2019.

\bibitem{chu2019fairdarts}
Xiangxiang Chu, Tianbao Zhou, Bo Zhang, and Jixiang Li.
\newblock Fair {DARTS:} eliminating unfair advantages in differentiable
  architecture search.
\newblock In {\em ECCV}, 2020.

\bibitem{DenseNAS}
Jiemin Fang, Yuzhu Sun, Qian Zhang, Yuan Li, Wenyu Liu, and Xinggang Wang.
\newblock Densely connected search space for more flexible neural architecture
  search.
\newblock In {\em CVPR}, 2020.

\bibitem{fang2020fna++}
Jiemin Fang, Yuzhu Sun, Qian Zhang, Kangjian Peng, Yuan Li, Wenyu Liu, and
  Xinggang Wang.
\newblock Fna++: Fast network adaptation via parameter remapping and
  architecture search.
\newblock {\em TPAMI}, 2020.

\bibitem{guo2020hit}
Jianyuan Guo, Kai Han, Yunhe Wang, Chao Zhang, Zhaohui Yang, Han Wu, Xinghao
  Chen, and Chang Xu.
\newblock Hit-detector: Hierarchical trinity architecture search for object
  detection.
\newblock In {\em CVPR}, 2020.

\bibitem{SPOS}
Zichao Guo, Xiangyu Zhang, Haoyuan Mu, Wen Heng, Zechun Liu, Yichen Wei, and
  Jian Sun.
\newblock Single path one-shot neural architecture search with uniform
  sampling.
\newblock In {\em ECCV}, 2020.

\bibitem{mobilenetv3}
GAndrew Howard, Mark Sandler, Grace Chu, Liang-Chieh Chen, Bo Chen, Mingxing
  Tan, Weijun Wang, Yukun Zhu, Ruoming Pang, Vijay Vasudevan, Quoc~V. Le, and
  Hartwig Adam.
\newblock Searching for mobilenetv3.
\newblock In {\em ICCV}, 2019.

\bibitem{StacNAS}
Guilin Li, Xing Zhang, Zitong Wang, Zhenguo Li, and Tong Zhang.
\newblock Stacnas: Towards stable and consistent optimization for
  differentiable neural architecture search.
\newblock {\em CoRR}, abs/1909.11926, 2019.

\bibitem{DARTS+}
Hanwen Liang, Shifeng Zhang, Jiacheng Sun, Xingqiu He, Weiran Huang, Kechen
  Zhuang, and Zhenguo Li.
\newblock {DARTS+:} improved differentiable architecture search with early
  stopping.
\newblock {\em CoRR}, abs/1909.06035, 2019.

\bibitem{AutoDeeplab}
Chenxi Liu, Liang{-}Chieh Chen, Florian Schroff, Hartwig Adam, Wei Hua, Alan~L.
  Yuille, and Fei{-}Fei Li.
\newblock Auto-deeplab: Hierarchical neural architecture search for semantic
  image segmentation.
\newblock In {\em CVPR}, 2019.

\bibitem{liu2017hierarchical}
Hanxiao Liu, Karen Simonyan, Oriol Vinyals, Chrisantha Fernando, and Koray
  Kavukcuoglu.
\newblock Hierarchical representations for efficient architecture search.
\newblock In {\em ICLR}, 2018.

\bibitem{liu2018darts}
Hanxiao Liu, Karen Simonyan, and Yiming Yang.
\newblock {DARTS}: Differentiable architecture search.
\newblock In {\em ICLR}, 2019.

\bibitem{ENAS}
Hieu Pham, Melody~Y. Guan, Barret Zoph, Quoc~V. Le, and Jeff Dean.
\newblock Efficient neural architecture search via parameter sharing.
\newblock In {\em ICML}, 2018.

\bibitem{AmoebaNet}
Esteban Real, Alok Aggarwal, Yanping Huang, and Quoc~V. Le.
\newblock Regularized evolution for image classifier architecture search.
\newblock In {\em AAAI}, 2019.

\bibitem{Cellbased}
Yao Shu, Wei Wang, and Shaofeng Cai.
\newblock Understanding architectures learnt by cell-based neural architecture
  search.
\newblock In {\em ICLR}, 2020.

\bibitem{DAAS}
Yunjie Tian, Chang Liu, Lingxi Xie, Jianbin Jiao, and Qixiang Ye.
\newblock Discretization-aware architecture search.
\newblock {\em CoRR}, abs/2007.03154, 2020.

\bibitem{wu2019fbnet}
Bichen Wu, Xiaoliang Dai, Peizhao Zhang, Yanghan Wang, Fei Sun, Yiming Wu,
  Yuandong Tian, Peter Vajda, Yangqing Jia, and Kurt Keutzer.
\newblock Fbnet: Hardware-aware efficient convnet design via differentiable
  neural architecture search.
\newblock In {\em CVPR}, 2019.

\bibitem{xu2020pcdarts}
Yuhui Xu, Lingxi Xie, Xiaopeng Zhang, Xin Chen, Guo{-}Jun Qi, Qi Tian, and
  Hongkai Xiong.
\newblock {PC-DARTS:} partial channel connections for memory-efficient
  architecture search.
\newblock In {\em ICLR}, 2020.

\bibitem{RobustDARTS}
Arber Zela, Thomas Elsken, Tonmoy Saikia, Yassine Marrakchi, Thomas Brox, and
  Frank Hutter.
\newblock Understanding and robustifying differentiable architecture search.
\newblock In {\em ICLR}, 2020.

\bibitem{zhang2019customizable}
Yiheng Zhang, Zhaofan Qiu, Jingen Liu, Ting Yao, Dong Liu, and Tao Mei.
\newblock Customizable architecture search for semantic segmentation.
\newblock In {\em CVPR}, 2019.

\bibitem{BayesNAS}
Hongpeng Zhou, Minghao Yang, Jun Wang, and Wei Pan.
\newblock Bayesnas: {A} bayesian approach for neural architecture search.
\newblock In {\em ICML}, 2019.

\bibitem{NASNet}
Barret Zoph and Quoc~V. Le.
\newblock Neural architecture search with reinforcement learning.
\newblock In {\em ICLR}, 2017.

\end{thebibliography}
}

\end{document}